\documentclass[]{spie}

\usepackage{amsmath,amsfonts,amssymb}
\usepackage{siunitx}
\usepackage[linesnumbered]{algorithm2e}
\usepackage{graphicx}
\usepackage{float}
\usepackage{tabularx}
\usepackage{multicol}
\usepackage{pgfplots}
\usepackage{subfigure}
\usetikzlibrary{shapes.arrows, arrows.meta, arrows, calc, shapes.misc, positioning, decorations.pathmorphing}
\pgfplotsset{every axis/.append style={tick label style={/pgf/number format/fixed},font=\scriptsize,ylabel near ticks,xlabel near ticks}}
\usepackage{tikz}
\usepackage{booktabs}
\pgfplotsset{compat=1.17}
\usepackage[colorlinks=true, allcolors=blue]{hyperref}
\usepackage[misc]{ifsym} 
\usepackage{color, colortbl}

\definecolor{LightCyan}{rgb}{0.9254902 , 0.67058824, 0.81568627}
\definecolor{blue}{rgb}{0, 0, 1}

\title{Histopathological Image Classification based on Self-Supervised Vision Transformer and Weak Labels}

\author{Ahmet Gokberk Gul\textsuperscript{*}}
\author{Oezdemir Cetin\textsuperscript{*}}
\author{Christoph Reich}
\author[2]{Nadine Flinner}
\author{Tim Prangemeier}
\author{Heinz Koeppl \textsuperscript{(\Letter)}}

\affil[1]{Self-Organizing Systems Lab, Department of Electrical Engineering and Information
Technology, Technische Universität Darmstadt, Germany}
\affil[2]{Dr. Senckenberg Institute of Pathology, University Hospital Frankfurt, Germany}

\authorinfo{\textsuperscript{*}Ahmet Gokberk Gul and Oezdemir Cetin — both authors contributed equally \\E-mail: \href{mailto:heinz.koeppl@tu-darmstadt.de}{\color{black}{heinz.koeppl@tu-darmstadt.de}}}

\begin{document} 
\maketitle

\begin{abstract}
Whole Slide Image (WSI) analysis is a powerful method to facilitate the diagnosis of cancer in tissue samples. Automating this diagnosis poses various issues, most notably caused by the immense image resolution and limited annotations. WSI's commonly exhibit resolutions of $100,000 \times 100,000$ pixels. Annotating cancerous areas in WSI's on the pixel-level is prohibitively labor-intensive and requires a high level of expert knowledge. Multiple instance learning (MIL) alleviates the need for expensive pixel-level annotations. In MIL, learning is performed on slide-level labels, in which a pathologist provides information about whether a slide includes cancerous tissue. Here, we propose Self-ViT-MIL, a novel approach for classifying and localizing cancerous areas based on slide-level annotations, eliminating the need for pixel-wise annotated training data. Self-ViT-MIL is pre-trained in a self-supervised setting to learn rich feature representation without relying on any labels. The recent Vision Transformer (ViT) architecture builds the feature extractor of Self-ViT-MIL. For localizing cancerous regions, a MIL aggregator with global attention is utilized. To the best of our knowledge, Self-ViT-MIL is the first approach to introduce self-supervised ViT's in MIL-based WSI analysis tasks. We showcase the effectiveness of our approach on the common Camelyon16 dataset. Self-ViT-MIL surpasses existing state-of-the-art MIL-based approaches in terms of accuracy and area under the curve (AUC). Code is available at \url{https://github.com/gokberkgul/self-learning-transformer-mil}
\end{abstract}

\keywords{Multiple Instance Learning, Weakly Supervised Learning, Vision Transformer, Computational Pathology, Cancer Detection, Whole Slide Image}

\section{Introduction}
\label{sec:intro} 

Histopathological whole slide scanning is a common and widely used tool for visualizing tissue samples \cite{pantanowitz_farahani_parwani_2015}. These visualizations are used for diagnosing diseases, such as cancer \cite{Bandi, Ehteshami}. Tissue samples are scanned by specialized microscopes on glass slides, resulting in digital whole slide images \cite{pantanowitz_farahani_parwani_2015}. Automating the diagnosis of diseases based on WSIs is a current challenge in computed aided diagnostics \cite{ilse2018attentionbased, courtiol2020classification, li2021dualstream, shao2021transmil}. WSI's exhibit extremely high resolutions of up to $100,000\times 100,000$ pixels \cite{pantanowitz_farahani_parwani_2015, Ehteshami}. This resolution makes it infeasible to directly process whole slides with common computer vision models, such as convolutional neural networks (CNNs) \cite{li2021dualstream, Jin2022}. To cope with such high image resolutions, WSIs are typically processed patch-wise, making the usage of CNNs feasible \cite{li2021dualstream, shao2021transmil}. Training deep networks typically requires a sophisticated amount of ground truth annotations \cite{LeCun2015}. Obtaining pixel-wise ground truth annotations is generally labor-intensive and time-consuming, for example, labeling a single $1024\times 2048$ image of the cityscapes dataset (natural urban scenes) is reported to have taken approximately $90\si{\min}$ \cite{Cordts2016}. In contrast to natural scenes, annotating WSIs requires a high level of expert knowledge, and pixel-wise annotations can differ between experts \cite{Bandi, Ehteshami, kohl2019probabilistic}. Motivated by these issues, we propose an approach for computed aided diagnostics based only on slide-level labels (weak labels).

Recent approaches for learning from weak WSI annotations follow the multiple instance learning (MIL) paradigm \cite{Carbonneau_2018, ilse2018attentionbased, li2021dualstream, shao2021transmil, lerousseau2021weakly}. In MIL, each WSI is considered as a bag. Each bag contains multiple instances. These instances are patches obtained by cropping the high-resolution WSI, resulting in low-resolution images. Labels only contain global information whether a bag contains disease-positive instances or not. Recent MIL-based approaches first extract patch-level latent representations with a CNN \cite{li2021dualstream, shao2021transmil}, then all latent representations are aggregated before being fed into a classifier \cite{li2021dualstream}. Training is typically performed in an end-to-end manner.

In recent years, CNNs have become the \textit{de facto} standard for medical image analysis, examples including cell segmentation \cite{ronneberger2015unet, prangemeier2020a, prangemeier2022}, WSI analysis \cite{li2021dualstream, Jin2022}, microscopy image synthesis \cite{Reich2021a}, and brain tumor segmentation \cite{Reich2021b}. More recently, transformer-based models \cite{Vaswani2017, prangemeier2020b, shao2021transmil}, such as the Vision Transformer (ViT) \cite{dosovitskiy2021image} have been proposed. These transformer-based models have shown satisfying performance over CNNs when trained in a self-supervised fashion \cite{dino}. Here we propose, Self-ViT-MIL that combines self-supervised learning, ViTs, and MIL. Self-ViT-MIL is evaluated on the common Camelyon16 dataset \cite{Ehteshami} against recent weakly supervised state-of-the-art approaches. Our approach outperforms all other approaches in terms of accuracy and AUC. Self-ViT-MIL also achieves competitive results against a fully-supervised baseline.

\section{Methods} \label{methods}

Self-ViT-MIL (Fig. \ref{alg:mil}) can be divided into two stages. In the first stage, the ViT is trained in a self-supervised fashion to learn rich feature representations of the input patches \cite{dino}. In the second stage, the ViT weights are frozen and a MIL aggregator is trained on top of the ViT's latent features \cite{li2021dualstream}.

\begin{figure}[H]
    \centering
    \includegraphics[width = \linewidth]{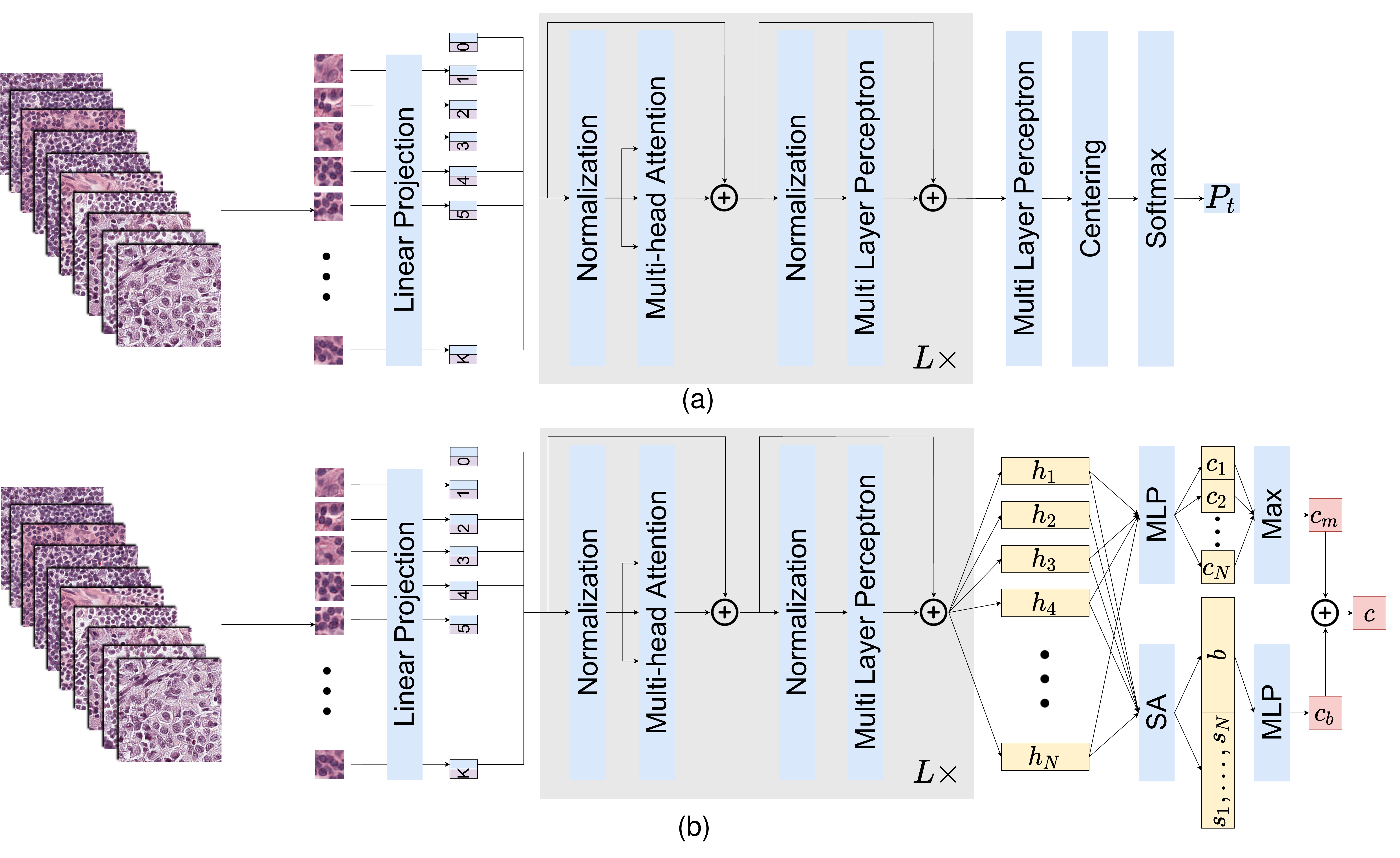}
    \caption{Self-ViT-MIL method flow chart. (a) Self-supervised training with DINO training approach. Although DINO \cite{dino} follows student-teacher approach, only the teacher network is shown for simplicity. Each patch extracted from the WSIs are further divided into $16 \times 16$ patches, flattened with linear projection layer and positional encodings are added. Encoded patches are passed through $L$ transformer encoder layers, which consists of normalization, multi-head attention, and MLP layers with residual connections. The transformer encoder is followed by an MLP, centering (only applied to the teacher network) and softmax layers. (b) Training of MIL aggregator. The trained ViT is used to generate feature embeddings of WSIs. The first stream calculates instance scores and maximum instance score. The second stream uses global self-attention the calculate a bag vector and attention scores, later used for localization. The bag vector is classified to estimate the bag score. The addition of the maximum instance score and the bag score builds the final score.}
    \label{fig:flow}
\end{figure}

\subsection{Self-Supervised Vision Transformer} \label{vit}

The Vision Transformer, proposed by Dosovitskiy \textit{et al.} is a general backbone for vision tasks \cite{dosovitskiy2021image}. The core building block of ViTs is an attention-based Transformer encoder, originally proposed in the domain of natural language processing \cite{Vaswani2017}. Let $\mathbf{X} \in \mathbb{R}^{C \times H \times W}$ be an input image with $C$, $H$ and $W$ as channel size, height, and width, respectively. $\mathbf{X}$ is partitioned into non-overlapping patches to generate the tensor $\mathbf{X}_p \in \mathbb{R}^{N \times P^2C}$ with $N = \frac{HW}{P^2}$. A linear layer of input size $P^2 C$ and output size of $D$ is used for the projection of the patches, which generates an $N \times D$ tensor. Class, \texttt{[cls]}, token of size $1 \times D$ is concatenated to the projected patches, making the tensor size $(N+1) \times D$. Transformers are agnostic to the permutation of the input tokens. To incorporate positional information, learnable positional encodings are employed. For each projected patch and the \texttt{[cls]} token, $1 \times D$ learnable positional embeddings are added. ViT consists of $L$ transformer encoder blocks. Each block contains a Multi-Head Self-Attention (MSA) layer, two Layer Normalizations, and a Multilayer Perceptron (MLP). The \texttt{[cls]} token at the output of the last transformer is used as a latent representation of the original image.

DINO is a self-supervised learning technique \cite{dino}. In the knowledge distillation paradigm, there are two networks, called student and teacher, with network notation and parameters $\{g_{\boldsymbol{\theta}_s}, \boldsymbol{\theta}_s\}$ and $\{g_{\boldsymbol{\theta}_t}, \boldsymbol{\theta}_t\}$, respectively. Loss is the cross-entropy function between student and teacher network output probability distribution. Student parameters $\boldsymbol{\theta}_s$ are updated with respect to the cross-entropy loss function. For each image $\mathbf{X}$, two global and several local views are generated, by sampling random transformations. All views of the image are stored in a set $V = \{ \Tilde{\mathbf{X}}_1^g, \Tilde{\mathbf{X}}_2^g, \Tilde{\mathbf{X}}_1^l, \dots, \Tilde{\mathbf{X}}_n^l \}$. $V$ is passed through the student network, while a partition of the set that contains only the global views, ${V}^g$, is fed to the teacher network to encourage local to global correspondences.

Differing from the knowledge distillation paradigm, there is no teacher network given in prior. The teacher is built from the past iterations of the student network. Teacher network is updated with exponential moving average (EMA) update \cite{dino}. Since there are no contrastive loss or clustering constraints, DINO uses a centering and sharping operations in order to avoid collapsing \cite{dino}. The centering operation avoids that a single dimension dominates the probability distribution, instead, it pushes the model to generate a uniform distribution. The sharpening operation has the opposite effect, so a joint usage of both operations crates a balance between uniform and single dimension domination, hence avoiding collapse.

\subsection{MIL Aggregator} \label{mil}

In the MIL setting, training data is arranged in sets, called bags, which contain instances. Class labels are only provided for the bags and instance labels are not known \textit{a priori}. \cite{Carbonneau_2018}. Prediction can be performed at the bag level, instance level, or both. Let $\mathcal{X}$ be a set, $\mathcal{X} = \{\mathbf{X_1}, \mathbf{X_2}, \dots, \mathbf{X_n} \} \: \text{where} \: \mathbf{X_i} \in \mathbb{R}^{C \times H \times W}, i \in \{1, \dots, n\}$. The standard assumption of MIL is that if any of the instance label is $1$, the bag label $Y$ is also $1$, otherwise $Y$ is also $0$. This approach is used in WSI classification tasks since image patches extracted from the WSI can be treated as instances. 

In Dual Stream Multiple Instance Learning (DSMIL) \cite{li2021dualstream} design, there are two streams, first stream learns the instance classifier, the most important instance and its score via max pooling with the formula $c_m(\mathcal{X}) = \text{max}\{ \mathbf{W}_p \mathbf{h}_1, \dots,  \mathbf{W}_p \mathbf{h}_N \}$. Each embedding $\mathbf{h}_i \in \mathbb{R}^{D \times 1}$, generated by the pre-trained ViT, is projected with a learnable matrix $\mathbf{W}_p \in \mathbb{R}^{1 \times K}$, creating a score for each class. 

Second stream aggregates the instances to obtain a embedded space representation of the bag, with a self attention module. From each embedding, query and information vectors are generated with learnable projection matrices $\mathbf{W}_q \in \mathbb{R}^{L \times K}$ and $\mathbf{W}_v \in \mathbb{R}^{K \times K}$. Distance between each query vector $\mathbf{q}_i$ to the query vector that corresponds to the critical instance $\mathbf{q}_m$ is calculated, shown in equation \ref{eqn:distance}.
\begin{equation}
\label{eqn:distance}
    s_i = D(\mathbf{h}_i, \mathbf{h}_m) = \frac{\exp(\langle \mathbf{q}_i, \mathbf{q}_m \rangle)}{ \sum_{k = 1}^N \exp(\langle \mathbf{q}_k, \mathbf{q}_m \rangle)  }  
\end{equation}
The resulting scalar, $s_i$, can be interpreted as the score of instance $\mathbf{h}_i$. These scores are later used to generate an attention map and localize the cancerous areas. Embedded space representation of the bag is achieved by the weighted sum of information vectors, with the distance measure as weights. Bag score is calculated with a learnable projection matrix $ \mathbf{W}_b \in \mathbb{R}^{1 \times K}$. The final score is the average of instance space and embedded space bag score. A pseudo algorithm can be found in Appendix \ref{sec:milalgo}.

\section{Experiments}

Our experiments utilize the common Camelyon16 dataset, consisting of $399$ WSIs \cite{Ehteshami}. The training set includes $110$ slides with nodal metastasis and $160$ without. The test dataset consists of $129$ WSIs, of which $49$ contain metatases. All annotations were generated by pathologists, however they are not used for training.

All WSIs are cropped to non-overlapping $224 \times 224$ patches on $20 \times$ magnification factor. For background filtering, the Otsu Thresholding \cite{Otsu} method is used. After filtering, the dataset consisted of $1,235,428$ training patches and $544,420$ testing patches. Stain normalization is applied during pre-processing to obtain more uniformly distributed patches \cite{Macenko}.

A ViT-B/16 is employed as the backbone network for feature extraction \cite{dosovitskiy2021image}. EMA update factor $\lambda$ is chosen to be $0.9995$ and ramps up to $1$ with a cosine scheduler during the training. In the warm-up phase, the learning rate linearly increases from $\eta_{min}$ to $\eta_{max}$. After the warm-up phase, the learning rate drops from $\eta_{max}$ to $\eta_{min}$, with a cosine learning rate scheduler. Choice for $\eta_{min}$ and $\eta_{max}$ were $10^{-6}$ and $5 \cdot 10^{-4}$, respectively. Similarly, teacher temperature $\tau_t$ has two phases. For the warm-up phase, the value is chosen to be $0.01$, which accelerates the loss decrease, but it changes to $0.04$ after the warm-up phase for robustness. ViT is trained on 4 Nvidia V100 GPUs (32GB) for 100 epochs with 10 epochs as the warm-up phase, each epoch took 55 minutes.

The pre-trained ViT weights are frozen during MIL aggregator training. The learning rate $\eta$ was $2 \cdot 10^{-5}$. Instead of performing the classification on the \texttt{[cls]} token of the last block,  the \texttt{[cls]} tokens of the last 4 blocks were stacked and their is average appended to the stack, generating a vector of $\mathbb{R}^{1 \times 5D}$ per patch. The MIL aggregator was trained using AdamW \cite{adamw} on a Nvidia V100 GPU (32GB) for 50 epochs ($\approx$ 27 minutes each).

\section{Results}

\begin{table}[H]
	\begin{minipage}{0.495 \linewidth}
        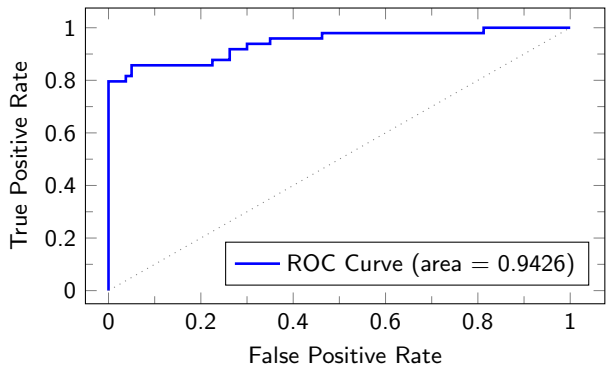
\begin{figure}[H]
            \centering
            \pgfplotstableread{images/dsmil_auc.dat}{\auc}
            \begin{tikzpicture}[every node/.style={font=\fontsize{9}{9}\sffamily}, >={Stealth[inset=0pt,length=6.0pt,angle'=45]}]
                \begin{axis}[
                    xmin = -0.05, xmax = 1.075,
                    ymin = -0.05, ymax = 1.075,
                    xtick distance = 0.2,
                    ytick distance = 0.2,
                    minor tick num = 1,
                    width = 1\textwidth,
                    height = 0.65\textwidth,
                    xtick={0, 0.2, 0.4, 0.6, 0.8, 1},
                    xticklabel={\fontsize{9}{9}\sffamily\pgfmathprintnumber[assume math mode=true]{\tick}},
                    ytick={0, 0.2, 0.4, 0.6, 0.8, 1},
                    yticklabel={\fontsize{9}{9}\sffamily\pgfmathprintnumber[assume math mode=true]{\tick}},
                    legend cell align = {left},
                    legend style={at={(0.97,0.21)}},
                    xlabel = {False Positive Rate},
                    ylabel = {True Positive Rate}
                ]
                \addplot[blue,line width=0.35mm] table [x = {fpr}, y = {tpr}] {\auc};
                \addplot[domain=0:1,gray,dotted, width=0.25mm] {x};
                \legend{
                    ROC Curve (area = 0.9426)
                }
            \end{axis}
            \end{tikzpicture}
            \caption{MIL Aggregator ROC Curve.}
            \label{fig:mil_acc}
        \end{figure}
	\end{minipage}\hfill
	\begin{minipage}{0.50\linewidth}
    	\centering
    	\caption{Numerical results of Self-ViT-MIL on the Camelyon16 dataset \cite{Ehteshami} compared with other methods.}
        \begin{tabular}{l c c c c @{}} \toprule
            Model & Scale & {Accuracy} & {AUC} \\ \midrule
            CHOWDER \cite{courtiol2020classification} & Single & {-} & 0.8580 \\
            ABMIL\textsuperscript{$\ddag$} \cite{ilse2018attentionbased} & Single & 0.8450 & 0.8653 \\
            DSMIL \cite{li2021dualstream} & Single & 0.8682  & 0.8944 \\
            MIL-RNN \cite{Campanella} & Single & {-}  & 0.8990 \\
            DSMIL-LC \cite{li2021dualstream} & Multiple & 0.8992 & 0.9165 \\ 
            TransMIL \cite{shao2021transmil} & Single & 0.8837 & 0.9309 \\
            Fully-supervised \cite{li2021dualstream} & Single & 0.9147 &  0.9362 \\
            \rowcolor{LightCyan}
            Self-ViT-MIL (ours) & Single & 0.9147 & 0.9426\\ 
            \bottomrule
            \multicolumn{3}{l}{\textsuperscript{$\ddag$}ABMIL results taken from DSMIL paper. \cite{li2021dualstream}}\\
        \end{tabular}
        \label{tab:results_compare}
	\end{minipage}
\end{table}

Our accuracy and AUC scores on the Camelyon16 dataset \cite{Ehteshami}, compared to other MIL-based approaches from the literature, are presented in Table \ref{tab:results_compare}. We also compare Self-ViT-MIL against a naive fully-supervised baseline with a ResNet18 backbone from the literature \cite{li2021dualstream}. All works have trained their model on Camelyon16 training slides, expect MIL-RNN \cite{Campanella}. MIL-RNN is trained on the MSK axillary lymph node dataset, which consist of $9,894$ WSIs, which is about $36$ times larger than Camelyon16 training dataset. Evaluation is performed again on the Camelyon16 test dataset. DSMIL-LC is the multiscale version of DSMIL approach, combining $20 \times$ and $5 \times$ magnification slides to extract both local and global features \cite{li2021dualstream}. Accuracy score of $0.9147$ and AUC score of $0.9426$ is achieved with Self-ViT-MIL method. The receiver operating characteristics (ROC) curve is given in Figure \ref{fig:mil_acc}. This would put Self-ViT-MIL method in the first place Camelyon16 challenge and fourth place in open on leaderboard in Whole-slide-image classification task. Results achieved surpasses other papers that reported Camelyon16 WSI classification scores following MIL approach, making Self-ViT-MIL the state of the art method.

\begin{figure}[H]
    \centering
    \includegraphics[width = \linewidth]{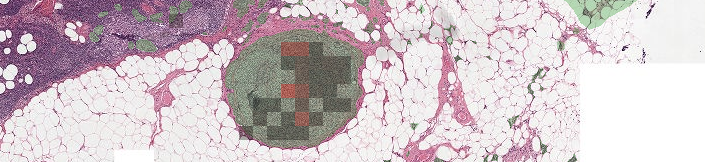}
    \caption{Camelyon16 Dataset Slide Test082, Zoomed in Localization Performance. Green area on the tissues shows the pathologists annotation. Transparent-black tiles show the model's prediction of the malignant patches. The amount of red on the predicted patches represents the confidence of the model.}
    \label{fig:test_082_zoomed}
\end{figure}

Localization performance is qualitatively presented in Figure \ref{fig:test_082_zoomed} (zoomed out version of the same slide can be seen in Figure \ref{fig:test_082}). Before drawing the model's prediction on the slide, Z-test is performed on attention scores to remove outliers and set 0 for the highest ones. DSMIL model is tested on the same slides, using the model weights given by the authors. DSMIL has a better coverage for the cancerous areas, but it also performs false positive prediction. Self-ViT-MIL model cannot cover all tumor areas, but almost no false positive predictions have occurred out of four random test slides selected.

\section{Discussion and Conclusion}

Our method outperforms both TransMIL \cite{shao2021transmil} and DSMIL-LC \cite{li2021dualstream}, the recent state-of-the-art MIL-based approaches, in terms of accuracy and AUC in the task of slide level classification on the Camelyon16 dataset. Self-ViT-MIL also achieved on par results with the DSMILs fully-supervised baseline \cite{li2021dualstream}. However, other fully-supervised methods with better performances have been reported.\footnote{\url{https://camelyon16.grand-challenge.org/Results/}} Training the ViT backbone in a self-supervised fashion is compute-intensive and requires a sophisticated compute infrastructure. Utilizing multi-scale input data showed performance improvements when applied to the DSMIL approach \cite{li2021dualstream}. Incorporating multi-scale input data to Self-ViT-MIL may also result in increased performance.

This paper presents Self-ViT-MIL, a model for weakly-supervised WSI classification. To the best of our knowledge, Self-ViT-MIL is the first approach to explore self-supervised ViTs in the application of WSI analysis. Self-ViT-MIL outperforms existing state-of-the-art weakly-supervised approaches on the Camelyon16 dataset.

\acknowledgments   
 
Calculations were conducted on the Lichtenberg high performance computer of the TU Darmstadt. H.K. acknowledges support from the European Research Council (ERC) with the consolidator grant CONSYN (nr. 773196). O.C. is supported by the Alexander von Humboldt Foundation Philipp Schwartz Initiative.

\bibliography{references} 
\bibliographystyle{spiebib} 

\appendix 

\section{Localization Results}
\label{sec:localization}

\begin{figure}[H]
    \centering
    \includegraphics[width = 0.87 \linewidth]{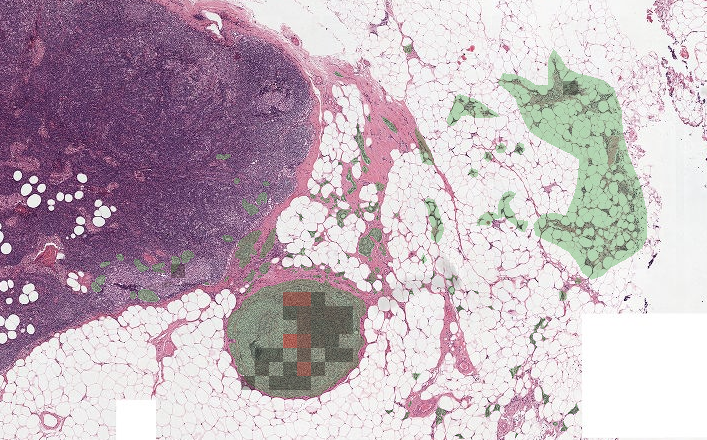}
    \includegraphics[width = 0.87 \linewidth]{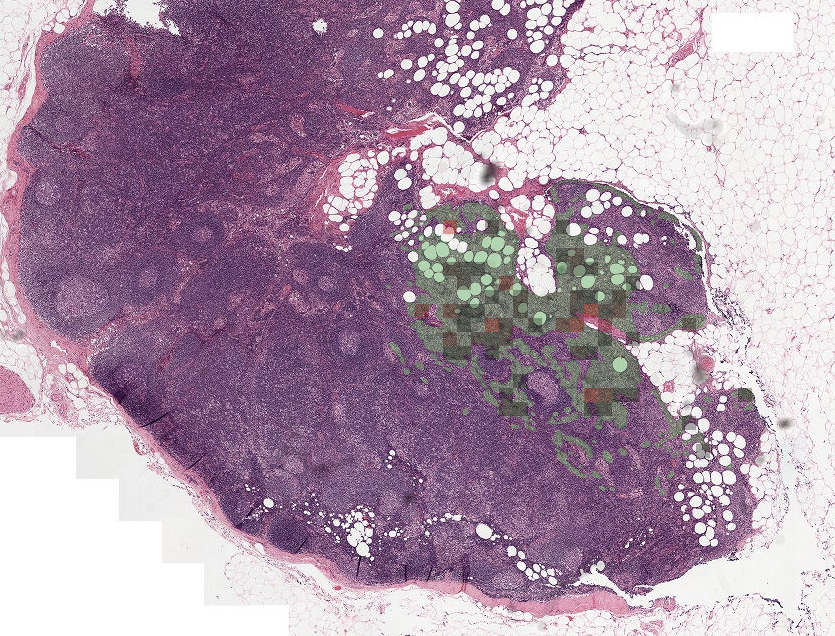}
    \caption{Camelyon16 Dataset Slide Test082, Localization Performance Around Tumor Areas.}
    \label{fig:test_082}
\end{figure}

\section{MIL Aggregator Algorithm}
\label{sec:milalgo}

\RestyleAlgo{ruled}
\begin{algorithm}
\DontPrintSemicolon
\caption{MIL Aggregator Algorithm}\label{alg:mil}
\ForEach {$ \{ \mathcal{X}, y \} \in \{\mathcal{X}^B, y^B \} $}{ 
    $\mathbf{h}_1, \dots, \mathbf{h}_N \gets f(\mathbf{X}_1), \dots, f(\mathbf{X}_N)$ \tcp*{Generate latent representation of image patches}
    
    $c_m(\mathcal{X}), \mathbf{h}_m \gets$ max$\{ \mathbf{W}_p \mathbf{h}_1, \dots,  \mathbf{W}_p \mathbf{h}_N \}$ \tcp*{Calculate instance score and critical instance}
    
    $\mathbf{q}_1, \dots, \mathbf{q}_N \gets \mathbf{W}_q \mathbf{h}_1, \dots, \mathbf{W}_q \mathbf{h}_N$ \tcp*{Generate query vectors}
    
    $\mathbf{v}_1, \dots, \mathbf{v}_N \gets \mathbf{W}_v \mathbf{h}_1, \dots, \mathbf{W}_v \mathbf{h}_N$ \tcp*{Generate information vectors}
    
    $s_1, \dots, s_N \gets D(\mathbf{h}_1, \mathbf{h}_m), \dots, D(\mathbf{h}_N, \mathbf{h}_m)$ \tcp*{Calculate instance scores}
    
    $\mathbf{b} \gets \sum_{i = 1}^N s_i \mathbf{v}_i$ \tcp*{Generate bag representation of the WSI}
    
    $c_b(\mathcal{X}) \gets \mathbf{W}_b \mathbf{b}$ \tcp*{Calculate bag score}

    $L \gets \frac{1}{2}(H(c_m(\mathcal{X}), y) + H(c_b(\mathcal{X}), y))$ \tcp*{Calculate loss score}
    
    $\mathbf{W}_p, \mathbf{W}_q, \mathbf{W}_v, \mathbf{W}_b \gets$ optimizer$(L, \mathbf{W}_p, \mathbf{W}_q, \mathbf{W}_v, \mathbf{W}_b, \eta)$ \tcp*{Update weights}
}
\end{algorithm}

\end{document}